# Vision-Based Lane-Changing Behavior Detection Using Deep Residual Neural Network

Zhensong Wei, Chao Wang, Peng Hao, *Member, IEEE*, and Matthew J. Barth, *Fellow, IEEE*

*Abstract*—Accurate lane localization and lane change detection are crucial in advanced driver assistance systems and autonomous driving systems for safer and more efficient trajectory planning. Conventional localization devices such as Global Positioning System only provide road-level resolution for car navigation, which is incompetent to assist in lane-level decision making. The state of art technique for lane localization is to use Light Detection and Ranging sensors to correct the global localization error and achieve centimeter-level accuracy, but the real-time implementation and popularization for LiDAR is still limited by its computational burden and current cost. As a cost-effective alternative, vision-based lane change detection has been highly regarded for affordable autonomous vehicles to support lane-level localization. A deep learning based computer vision system is developed to detect the lane change behavior using the images captured by a front-view camera mounted on the vehicle and data from the inertial measurement unit for highway driving. Testing results on real-world driving data have shown that the proposed method is robust with real-time working ability and could achieve around 87% lane change detection accuracy. Compared to the average human reaction to visual stimuli, the proposed computer vision system works 9 times faster, which makes it capable of helping make life-saving decisions in time.

## I. INTRODUCTION

Most autonomous driving systems perceive their surroundings through a wide variety of sensors, such as radar, Light Detection and Ranging (LiDAR), Global Positioning System (GPS), cameras and the inertial measurement unit (IMU) [1]. An internal map is then generated with collected inputs and guides the autonomous vehicle driving in a safe, fast and energy-saving manner [2]. Various techniques and applications have been developed in the fields of intelligent transportation systems to achieve different levels of vehicle automation, including speed assistance, collision avoidance, lane keeping, etc. [3]. Among them, the vehicle localization technique is one of the most important because an accurate positioning system allows the autonomous vehicle to understand its surrounding with less effort and operate the driving command safely. The most common vehicle localization method is to use GPS, which provides absolute positioning information at a low cost. However, GPS is vulnerable to various interference, such as tall buildings, trees or other signals which could prevent the GPS device from receiving satellite signals [4]. Moreover, the positioning error could reach up to tens of meters and therefore only road level resolution can be achieved [5]. Therefore, other methods are invented to fuse with GPS so that they can compensate for the large positioning error or used as back-up sensors when GPS is temporarily unavailable. For example, GPS - LiDAR fusion technique uses the LiDAR point cloud to estimate the incremental motion and model the error covariance in LiDAR-based position measurements, and the globally referenced pose is calculated using Unscented Kalman Filter given the measured error [6]. IMU enables the dead-reckoning method, but its error accumulation can cause errors for long term use [7]. Cameras, which usually act as visual odometers, can capture more information and are relatively cost-effective. By analyzing the lane-changing behavior of the vehicle through camera images and integrating the obtained information with a GPS, the tracking algorithm could tell the exact lane the vehicle is on and the localization accuracy can be enhanced up to centimeter [8]. The inexpensive vision-based localization method has drawn tremendous attention of researchers and a series of lane-changing detection applications have come up throughout the years to improve the detection accuracy.

While many early efforts on camera-based lane departure detection rely on lane boundary modeling [9-11], the overall image information is not considered comprehensively, making the system very sensitive to camera calibration and lack of potential to adapt complex conditions like high-density traffic. In [12], the authors employed a support vector machine (SVM) based framework to detect the lane-changing behavior using the edge information extracted from the pre-defined region of interest in the original image. The principal component analysis (PCA) was applied to reduce the dimension of the image features while keeping its energy, and the algorithm reached an accuracy of 68.5% when tested on actual driving data. A convolutional neural network (CNN) [13] based lane-change classifier was also implemented using the extracted edges as input and reached an accuracy of 79.7%. Higher detection accuracy was reached in [14], where the author applied a stacked sparse autoencoder model to classify the lane changing behaviors using the extracted useful features from images. A series of image preprocessing techniques were used to remove noise and enhance the classification accuracy, such as graying, filtering, binarization and setting a dynamic region of interest. An accuracy of 96.69% was achieved when testing on a total of 5309 frames of image sequences. A linear model was used for lane detection, therefore the algorithm might not work well for a larger scale of implementation with more complex road situations. In recent years, deep learning (DL) approaches are increasingly applied to vision-based lane departure detection for end-to-end solutions. In

Zhensong Wei, Chao Wang, Peng Hao, and Matthew J. Barth are with the College of Engineering – Center for Environmental Research and Technology (CE-CERT), University of California, Riverside, CA 92507, USA. Email: zwei030@ucr.edu, cwang061@ucr.edu, haop@cert.ucr.edu, barth@ece.ucr.edu.

[15], lane positions were estimated using an end-to-end deep neural network. Lanes were carefully marked in the images captured by a laterally-mounted down-facing camera and used as labels to the network. The detected lanes could then be used for the lane changing detection or lane departure warning. Despite the various vision-based lane detection methods, there hasn't been an end-to-end lane-changing behavior detection system using captured images directly.

In this paper, we proposed a vision-based real-time system to detect the lane-changing behavior of the vehicle. We designed deep residual learning neural networks to recognize the images captured by a forward-facing, in-vehicle camera and determine the three lane-changing behaviors: lane departure to the left (class 1); to the right (class 2); or no lane departure (class 3). To further improve the accuracy, another network was designed to utilize the IMU information as additional input. A baseline model using only IMU data was also developed for comparison. The NUDrive 1000 lane-change dataset [16] was used to train, validate and test the proposed models. The results indicate the proposed method achieves good lane change detection accuracy and is capable to work in real time. The main contribution of this paper is to propose a deep learning based computer vision system for lane change behavior, characterized by these key novelties: 1) It offers a novel end-to-end solution directly from image to lane change detection and no other model (like lane marks detection model) in the middle is involved; 2) As one of the leading edge deep learning algorithms, to the authors' knowledge, ResNet has not been customized and integrated with lane change behavior detection yet; 3) We improve the performance of the image-input-only ResNet by novelly concatenating the IMU data with the feature map (the pooling layer) before the fully connected layer; 4) Our system effectively and efficiently detects the lane-changing behavior with 87% of accuracy and 0.028s of response time.

The remaining of the paper is organized as follows: Section II presents a detailed description of the proposed deep neural network; Section III shows the testing results with a comparison of the proposed models and the baseline model; Section IV concludes the paper and discusses the future work.

## II. METHODOLOGY

### A. Dataset Description and Data Preprocessing

In this work, the NUDrive 1000 lane change dataset [16] is applied to study the lane-changing behavior. The driving data was collected on real highways around Nagoya, Japan. Video images of the road in front of the vehicle were recorded by a front-facing camera, and vehicle operation signals including vehicle velocity, acceleration, and gas and brake pedal pressures were recorded using different sensors. Ten drivers were recruited to complete approximately 50 km of highway driving for each driver. The images were sampled from the driving videos at 10Hz and labeled with one of the three lane-changing behaviors, lane departure to the left, to the right, or no lane departure. Table 1 shows the three class labels and how they are represented in the dataset. The beginning and end of lane changes are defined as the time when lane markers appeared to begin moving or stop moving laterally on the front-view video respectively. Sample images representing the three classes are shown in Fig. 1 respectively. The dataset is large enough (around 200k training and 25k testing images) to prove the effectiveness of the algorithm when applied to real-world highway driving scenarios.

TABLE 1. Classes of Lane-changing Behavior.

| Class | Behavior | Class Representation |
|---|---|---|
| 1 | Lane departure to the left. | -1 |
| 2 | Lane departure to the right. | 1 |
| 3 | Keep in lane. | 0 |

As shown in Fig. 2, we crop out the upper and lower border of each image, where useful information is rarely contained. The smaller image has less noise and saves computational power for both training and testing. For each image, there is a set of corresponding IMU data which contains information of brake and gas pedal force, velocity, steering angle, and longitudinal and lateral acceleration of the vehicle. A detailed description of the IMU data is presented in Table 2. Those IMU data are combined as a 6-entry vector for the model input.

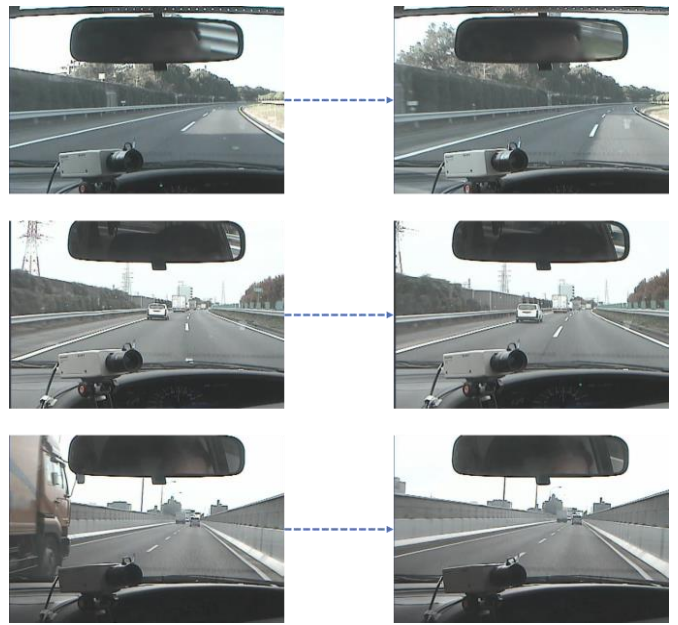

Figure 1. Sample images of the three lane-changing behaviors: lane departure to the left (upper); to the right (middle); or no lane departure (bottom).

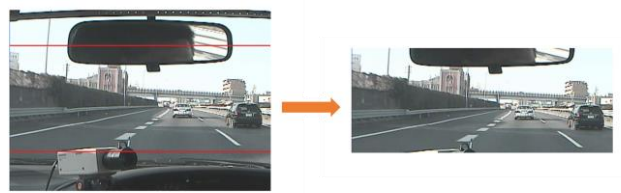

Figure 2. Image input is partially cropped to get rid of useless information. The cropped image has size 278×692 compared to the original image (480×692).

TABLE 2. Description of IMU Data.

| Signal | Description |
|---|---|
| Force on brake pedal [N] | Force on the brake pedal obtained by a pressure sensor. |
| Force on gas pedal [N] | Force on the gas pedal obtained by a pressure sensor. |
| Velocity [km/h] | Vehicle velocity obtained by a pulse generator. |
| Steering angle [deg.] | Steering wheel angle obtained by a potentiometer. |
| Longitudinal acceleration [G] | Straight line acceleration obtained by an accelerometer. |
| Lateral acceleration [G] | Centripetal acceleration obtained by an accelerometer. |

### B. Problem Description

The real-time lane change detection aims to understand the driving behavior and determine the current lane-changing status, which can be modeled as a classification problem:

$$y = f(x) \qquad (1)$$

where $f$ represents the classifier, $y \in \{-1, 0, 1\}$ is the output representing the class label, $x$ represents the input observation. The main objective of this study is to train a classifier that can take a frame of image (and/or IMU data) as input, and output the class of lane-changing state of the vehicle. As introduced above, we have two different types of observations: image and IMU. For each instance, the image is expressed by 3-channel 2D pixels and the IMU data is a 6-entry vector. To fully evaluate these two types of input, we develop separate models to test the detection accuracy for the 3 different input combinations, IMU only, image only, and image combined with IMU. Three models with different structures are trained to accommodate those inputs. The description of the method will be introduced in the following subsections.

### C. Network Architecture for Image Input Only

For the model with image input only, a deep neural network is trained using an adapted design of ResNet [17] with its architecture shown in the flow chart in Fig. 3. Resnet develops a residual learning framework which enables faster optimization with deeper network structure. This avoids a common problem named as "degradation" for "plain" net, in which simply stacked layers commonly result in higher training error when the network depth increases [18]. This network allows us to train with more layers and larger datasets at a faster speed (~ 200k images trained in 80 hours in our case).

The network receives three feature maps (RGB channels) as input, followed by the first convolutional layer (Conv) which expands the number of channels to 16. After the initial expansion, 12 residual blocks are connected after one another which increases the number of channels to 1024 and downsamples the feature map to 1/32 of its original size. The residual blocks would help improve the training speed of the network with its structure shown below:

$$Y = X + \operatorname{Re}LU(Conv(\operatorname{Re}LU(Conv(X)))) \qquad (2)$$

where ReLU is the Rectified Linear Unit used as the nonlinear activation function after convolutional layers, defined as ReLU(Z) = max(0, Z). X is zero padded to match the increasing dimension of the convolutional layers. After the residual blocks, an average pooling layer that eliminates the first two dimensions is used so that images of any size can be feed into the network without dimension mismatch. Two fully connected layers (FC) and a softmax layer (Softmax) are used in the end that output a 3-entry vector as an indication of the probability that image belongs to each of the three classes. The softmax function is defined as:

$$y_i = \frac{e^{z_i}}{\sum_{k=1}^{3} e^{z_k}} \quad \text{for i = 1, 2, 3} \qquad (3)$$

The label is set to be [1, 0, 0] for class 1, [0, 1, 0] for class 2 and [0, 0, 1] for class 3 and the network loss is defined as a cross entropy loss with equation shown below:

$$L = -\sum_i y_i \log x_i \qquad (4)$$

where $y_i$ and $x_i$ are the entry of label and network output respectively. The kernel size is set to be 3×3 with stride equals 1 for the convolutional layers and 2×2 with stride equals 2 for the max-pooling layers. During the training, the learnable parameters including weights, biases, and filters are updated using an adaptive moment estimation (Adam) [19] optimizer with a learning rate of 1×10$^{-4}$ and batch size of 16.

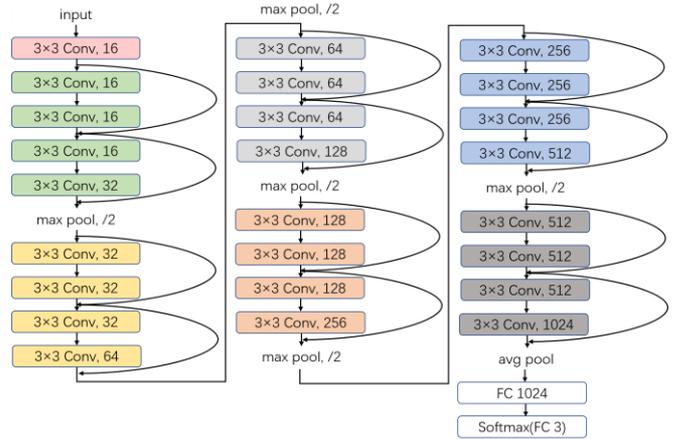

Figure 3. Network architecture trained for lane-changing detection using only images. The network contains 27 layers with around 1.1 million trainable parameters.

A summary of the output size and dimension of the convolutional filters at each layer is shown in Table 3. Conv_1 is the first convolutional layer and Conv_i_j is the jth layer in the ith block. A block contains two residual connections and is represented by the same color in Fig 3. Max pooling layer that downsamples the image by 4 is performed at Conv_2_4, Conv_3_4, Conv_4_4, Conv_5_4, and Conv_6_4, and an average pooling layer is performed after Conv_7_4 to transform the output size to 1024×1.

TABLE 3. Network Outputs and Convolutional Kernels Sizes.

| Layer Name | Output Size | Filter dimension |
|---|---|---|
| Conv_1 | 278×692 | [3×3, 16, stride 1] |
| Conv_2_x | 278×692 | [3×3, 16, stride 1] × 2 |
| | | [3×3, 16, stride 1] |
| | | [3×3, 32, stride 1] |
| Conv_3_x | 139×346 | [3×3, 32, stride 1] × 2 |
| | | [3×3, 32, stride 1] |
| | | [3×3, 64, stride 1] |
| Conv_4_x | 70×173 | [3×3, 64, stride 1] × 2 |
| | | [3×3, 64, stride 1] |
| | | [3×3, 128, stride 1] |
| Conv_5_x | 35×87 | [3×3, 128, stride 1] × 2 |
| | | [3×3, 128, stride 1] |
| | | [3×3, 256, stride 1] |
| Conv_6_x | 18×44 | [3×3, 256, stride 1] × 2 |
| | | [3×3, 256, stride 1] |
| | | [3×3, 512, stride 1] |
| Conv_7_x | 9×22 | [3×3, 512, stride 1] × 2 |
| | | [3×3, 512, stride 1] |
| | | [3×3, 1024, stride 1] |
| FC_1 | 1024×1 | |
| Softmax (FC_2) | 3×1 | |

The network is implemented using TensorFlow [20]. The training process takes ~400,000 iterations, which is about 34 epochs. All the model training, validation, and testing were all performed on a PC with four-core 4.20 GHz CPU, 64GB of RAM, and Nvidia GeForce GTX 1080 GPU.

### D. Network Architecture for Image and IMU Combined Input

We modified the previous network structure so that it could receive both types of input. We concatenated the one-dimensional IMU data with the average pooling layer in Fig. 3 before feeding them into the fully connected layer, as shown in Fig. 4, and kept all the other layers the same. The omitted structure before the blue arrow is exactly the same as the layout of the convolutional layers in Fig. 3, Also, the exact training parameters including learning rate, batch size, and loss function are set to be the same. This is to ensure that the structure and number of trainable parameters are close to the previous network and a fair comparison can be made to the detection accuracy with and without the additional IMU input.

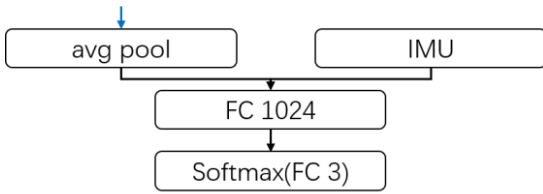

Figure 4. Concatenated IMU data with an average pooling layer

### E. Model for IMU Input Only

Using only IMU data to make lane change behavior predictions were well studied in the first Intelligent Transportation Systems plus Data Mining challenge during 2017 IEEE Intelligent Transportation Systems Conference [21]. There were twenty-three academy and industry contestants, coming from eleven different countries competed in the challenge and offered a variety of statistical and machine learning solutions. Among the solutions, the gradient boosting tree and the random forest are the top two models ranking with prediction accuracy. We reproduced the gradient boosting tree model as the baseline for the proposed vision-based methods.

Tree boosting is a highly effective and widely used machine learning method. Usually, a single tree is not strong enough to be used in practice. Tree boosting is an ensemble model which sums the prediction of multiple trees together. The method we used is a scalable end-to-end tree boosting system called XGBoost, which is used widely by data scientists to achieve state-of-the-art results on many machine learning challenges. More details of this method can be found in [22].

### III. RESULTS

All the three models are trained with the same dataset which contains 187,440 training images, 4500 validation images, and 24,626 testing images with each image of size 480 × 692. And for each image, there is a corresponding IMU containing 6-entry vector. The training input is fed into the model to train the parameters, the best model is chosen from the least loss or highest accuracy validated using the validation dataset, and the obtained model is tested with the testing dataset to calculate the accuracy. All the testing images are in different video clips from training and validation dataset to ensure the trained model is applicable to different road scenarios. Testing results for the three different models are shown in subsection A, B, and C.

### A. Testing Result for Training with IMU Only

In [21], the gradient boosting tree achieves 86.9% testing accuracy. However, considering the highly unbalanced testing data with 83.8% non-lane change rate, the accuracy is only 3% above a trivial guess.

In order to compare the baseline with proposed methods under the same condition, we applied the reproduced method to the same training, validation and testing set used for our proposed methods. Due to a 51.6% of non-lane change rate in our testing set, the testing accuracy is 55.6%, which is 4% above a trivial guess, indicating a similar performance as it was realized in [21]. A detailed test result is shown in Table 4.

TABLE 4. Testing Result for Tree Boosting Model Trained with IMU Data Only.

| Result | Class 1 | Class 2 | Class 3 | Total |
|---|---|---|---|---|
| Training Data | 18778 | 19178 | 149484 | 187440 |
| Validation Data | 1500 | 1500 | 1500 | 4500 |
| Testing Data | 5898 | 6033 | 12695 | 24626 |
| Testing Positive | 587 | 880 | 12216 | 13683 |
| Testing Negative | 5311 | 5153 | 479 | 10943 |
| Testing Accuracy | 9.95% | 14.59% | 96.23% | 55.56% |

## B. Testing Result for Training with Images Only

After the training of a deep CNN with structure from Fig. 3, the testing results are listed in Table 5. The results show that the proposed network structure is capable to identify the three categories of lane-changing behavior at an accuracy of 85.43%. The convergence plot is shown in Fig. 5. The training accuracy achieves 90% at around 35000 iterations.

TABLE 5. Testing Result for Training with Image Data Only

| Result | Class 1 | Class 2 | Class 3 | Total |
|---|---|---|---|---|
| Testing Data | 5898 | 6033 | 12695 | 24626 |
| Testing Positive | 5304 | 5093 | 10640 | 21037 |
| Testing Negative | 594 | 940 | 2055 | 3589 |
| Testing Accuracy | 89.93% | 84.42% | 83.81% | 85.43% |

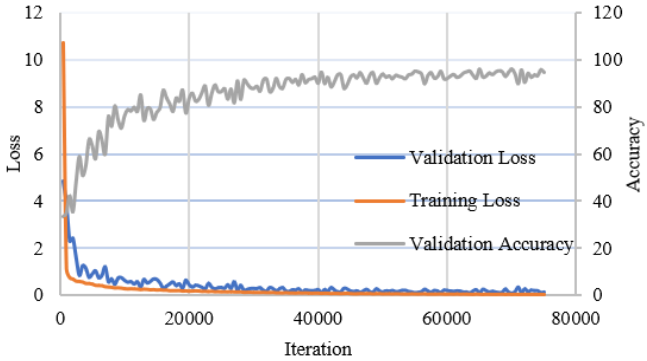

Figure 5. Convergence plot for the network trained with image only. The training takes a total of ~410k iterations, and 75k iterations are shown here since the convergence is too slow after 60k iterations.

## C. Testing result for Training with Image and IMU

After training of neural network with structure from Fig. 4, the testing results are listed in Table 6. The proposed network outperformed the previous two networks with an accuracy of 86.95% showing the network capability of utilizing two types of data. The convergence plot is shown in Fig 6. The training accuracy reached 90% at around 35000 iterations.

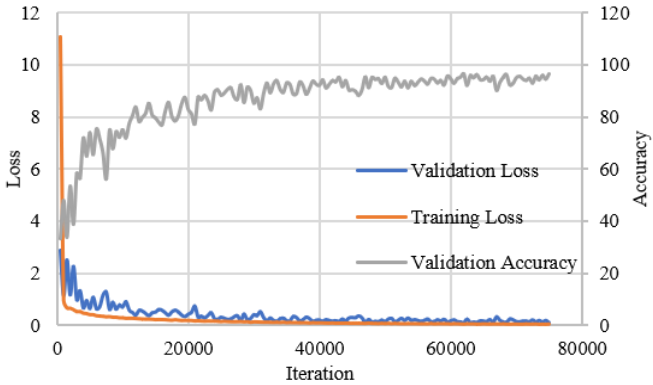

Figure 6. Convergence plot for the network trained with both image and IMU data.

TABLE 6. Testing Result for Training with Both Image and IMU Data

| Result | Class 1 | Class 2 | Class 3 | Total |
|---|---|---|---|---|
| Testing Data | 5898 | 6033 | 12695 | 24626 |
| Testing Positive | 5194 | 5350 | 10868 | 21412 |
| Testing Negative | 704 | 683 | 1827 | 3214 |
| Testing Accuracy | 88.06% | 88.68% | 85.61% | 86.95% |

## D. Comparison and Discussion

Speed of convergence comparison between network trained with image data only and images combined with IMU is shown in Fig. 7. The similar decreasing of losses shows that the two training processes have a similar speed of convergence. The training time, the number of iterations, testing time and accuracy for each method are listed in Table 7.

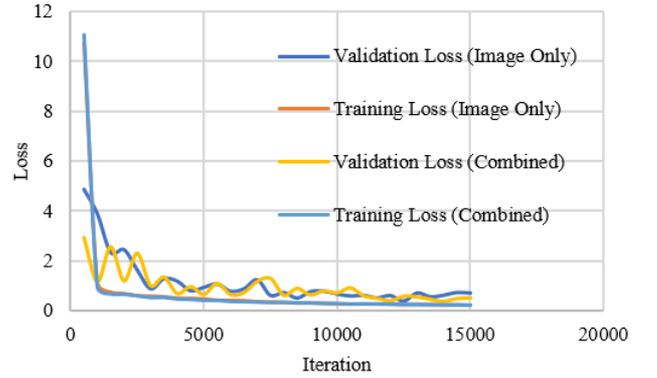

Figure 7. Speed of convergence comparison between the network trained with image data only and images combined with IMU. It shows that the two training processes have a similar speed of convergence.

TABLE 7. Training Time, Number of Iterations, Inference Time and Accuracy for Different Methods

| Model | Training time (s) | Iteration | Inference Time (s/image) | Accuracy |
|---|---|---|---|---|
| IMU Only | 101 | 8500 | $3.44 \times 10^{-7}$ | 55.56% |
| Image Only | 303733 | 412000 | 0.0276 | 85.43% |
| Image + IMU | 519724 | 647500 | 0.0278 | 86.95% |

The network trained with only IMU performs the worst because IMU data contains the least amount of information, only 6 values per time step. On the other hand, images contain way more information than IMU data and give a better result. The combination of the two achieves the best result. The inference time for a single image using the proposed models is below 0.028 seconds, which is capable of processing over 35 tuples of input per second. This means the proposed models reach the real-time level of computational speed. For a camera with a frame rate of 60, the lane change detection is able to respond for less than every two frames, which is quick enough for making life-saving decisions. In comparison, the average reaction time for humans is 0.25 seconds [23] to a visual stimulus. Our models are 9 times faster than human reaction.

## IV. CONCLUSION AND FUTURE WORK

This research explores end-to-end vision-based real-time detection approaches for identifying highway lane-changing behavior using deep learning. We designed deep residual learning neural networks to recognize the images captured by a forward-facing, in-vehicle camera and determine the three lane-changing behaviors: lane departure to the left (class 1); to the right (class 2); or no lane departure (class 3). To further improve the accuracy, another network was designed to utilize the IMU information as additional input. A baseline model using only IMU data was also developed for comparison. The NUDrive 1000 lane-change dataset was applied to train, validate and test the proposed models. The testing results on over 24k images show that the proposed method can achieve a detection accuracy of 86.95% on using both image and IMU data. The testing time of 0.0278 s/image also indicates the real-time working ability of the proposed method. Compared to the average human reaction to visual stimuli, the proposed computer vision system works 9 times faster, which makes it capable of helping make life-saving decisions in time. In the future, more research will be conducted as listed below:

• Extend the highway lane changing identification system to a local street where road scenarios are more complex.

• Since the data is in time sequences, we would like to test different recurrent neural network (RNN) structures, ex: Long short-term memory (LSTM), where the entire sequence of data is analyzed, and compare with the current network result.

ACKNOWLEDGMENT

The author would like to thank Dr. Chiyomi Miyajima and Dr. Kazuya Takeda at Nagoya University for kindly offering the dataset and assistance for this study.